\documentclass{article}

\usepackage{microtype}
\usepackage{graphicx}
\usepackage{subfigure}
\usepackage{booktabs} 

\usepackage[accepted,nohyperref]{icml2019}

\usepackage[utf8]{inputenc} 
\usepackage[T1]{fontenc}    
\usepackage{url}

\usepackage{amsmath,amsfonts,bm}

\def\eqref#1{equation~\ref{#1}}

\def\1{\bm{1}}

\DeclareMathAlphabet{\mathsfit}{\encodingdefault}{\sfdefault}{m}{sl}
\SetMathAlphabet{\mathsfit}{bold}{\encodingdefault}{\sfdefault}{bx}{n}

\newcommand{\Var}{\mathrm{Var}}

\newcommand{\Cov}{\mathrm{Cov}}

\usepackage{booktabs}       
\usepackage{amsthm}
\usepackage{amsfonts, amssymb}       
\usepackage{nicefrac}       
\usepackage{microtype}      
\usepackage{amsmath}
\usepackage{dsfont}
\usepackage{todonotes}

\newtheorem{theorem}{Theorem}[section]
\newtheorem{proposition}{Proposition}[section]

\icmltitlerunning{Efficient Optimization of Loops and Limits}

\begin{document}

\twocolumn[
\icmltitle{Efficient Optimization of Loops and Limits
with Randomized Telescoping Sums}




\begin{icmlauthorlist}
\icmlauthor{Alex Beatson}{pton}
\icmlauthor{Ryan P. Adams}{pton}
\end{icmlauthorlist}

\icmlaffiliation{pton}{Department of Computer Science, Princeton University, Princeton, NJ, USA}

\icmlcorrespondingauthor{Alex Beatson}{abeatson@cs.princeton.edu}

\icmlkeywords{Machine Learning, Optimization, Sequence, Iterative, Meta-learning, Meta, Learning, Meta-optimization, ODE, Differential Equations, Unbiased estimation}

\vskip 0.3in
]



\printAffiliationsAndNotice{}  

\begin{abstract}
  We consider optimization problems in which the objective requires an inner loop with many steps or is the limit of a sequence of increasingly costly approximations.
  Meta-learning, training recurrent neural networks, and optimization of the solutions to differential equations are all examples of optimization problems with this character.
  In such problems, it can be expensive to compute the objective function value and its gradient, but truncating the loop or using less accurate approximations can induce biases that damage the overall solution.
  We propose \emph{randomized telescope} (RT) gradient estimators, which represent the objective as the sum of a telescoping series and sample linear combinations of terms to provide cheap unbiased gradient estimates.
  We identify conditions under which RT estimators achieve optimization convergence rates independent of the length of the loop or the required accuracy of the approximation.
  We also derive a method for tuning RT estimators online to maximize a lower bound on the expected decrease in loss per unit of computation.
  We evaluate our adaptive RT estimators on a range of applications including meta-optimization of learning rates, variational inference of ODE parameters, and training an LSTM to model long sequences.
\end{abstract}

\section{Introduction}
Many important optimization problems consist of objective functions that can only be computed iteratively or as the limit of an approximation.
Machine learning and scientific computing provide many important examples.
In meta-learning, evaluation of the objective typically requires the training of a model, a case of bi-level optimization.
When training a model on sequential data or to make decisions over time, each learning step requires looping over time steps.
More broadly, in many scientific and engineering applications one wishes to optimize an objective that is defined as the limit of a sequence of approximations with both fidelity and computational cost increasing according to a natural number~${n\geq 1}$.
Inner-loop examples include: integration by Monte Carlo or quadrature with~$n$ evaluation points; solving ordinary differential equations (ODEs) with an Euler or Runge Kutta method with~$n$ steps and~$\mathcal{O}(\frac{1}{n})$ step size;
and solving partial differential equations (PDEs) with a finite element basis with size or order increasing with~$n$.

Whether the task is fitting parameters to data, identifying the parameters of a natural system, or optimizing the design of a mechanical part, in this work we seek to more rapidly solve problems in which the objective function demands a tradeoff between computational cost and accuracy.
We formalize this by considering parameters~$\theta\in\mathbb{R}^D$ and a loss function~$\mathcal{L}(\theta)$ that is the uniform limit of a sequence~$\mathcal{L}_n(\theta)$:
\begin{align}
\min_\theta \mathcal{L}(\theta) &= \min_\theta\lim_{n\to H}\mathcal{L}_n(\theta)\,.
\label{eqn:opt-problem}
\end{align}
Some problems may involve a finite horizon~${H}$, in other cases~${H = \infty}$.
We also introduce a cost function~${C:\mathbb{N}_{+}\to\mathbb{R}}$ that is nondecreasing in~$n$ to represent the cost of computing $\mathcal{L}_n$ and its gradient.

A principal challenge of optimization problems with the form in Eq.~\ref{eqn:opt-problem} is selecting a finite~$N$ such that the minimum of the surrogate~$\mathcal{L}_N$ is close to that of~$\mathcal{L}$, but without~$\mathcal{L}_N$ (or its gradients) being too expensive.
Choosing a large $N$ can be computationally prohibitive, while choosing a small~$N$ may bias optimization.
Meta-optimizing learning rates with truncated horizons can choose wrong hyperparameters by orders of magnitude \cite{wu2018understanding}.
Truncating backpropogation through time for recurrent neural networks (RNNs) favors short term dependencies \cite{tallec2017unbiasing}.
Using too coarse a discretization to solve an ODE or PDE can cause error in the solution and bias outer-loop optimization.
These optimization problems thus experience a sharp trade-off between efficient computation and bias.

We propose \emph{randomized telescope} (RT) gradient estimators, which provide cheap unbiased gradient estimates to allow efficient optimization of these objectives.
RT estimators represent the objective or its gradients as a telescoping series of differences between intermediate values, and draw weighted samples from this series to maintain unbiasedness while balancing variance and expected computation.

The paper proceeds as follows.
Section 2 introduces RT estimators and their history.
Section 3 formalizes RT estimators for optimization, and discusses related work in optimization.
Section 4 discusses conditions for finite variance and computation, and proves RT estimators can achieve optimization guarantees for loops and limits.
Section 5 discusses designing RT estimators by maximizing a bound on expected improvement per unit of computation.
Section 6 describes practical considerations for adapting RT estimators online.
Section 7 presents experimental results.
Section 8 discusses limitations and future work.
Appendix A presents algorithm pseudocode.
Appendix B presents proofs.
Code may be found at \url{https://github.com/PrincetonLIPS/randomized_telescopes}.


\section{Unbiased randomized truncation}
In this section, we discuss the general problem of estimating limits through randomized truncation.
The first subsection presents the randomized telescope family of unbiased estimators, while the second subsection describes their history (dating back to von Neumann and Ulam).
In the following sections, we will describe how this technique can be used to provide cheap unbiased gradient estimates and accelerate optimization for many problems.

\subsection{Randomized telescope estimators}
Consider estimating any quantity~${Y_H := \lim_{n \to H} Y_n}$ for~${n\in\mathbb{N}_{+}}$ where~${H\in\mathbb{N}_{+}\cup\{\infty\}}$.
Assume that we can compute $Y_n$ for any finite~${n\in\mathbb{N}_{+}}$, but since the cost is nondecreasing in~$n$ there is a point at which this becomes impractical.
Rather than truncating at a fixed value short of the limit, we may find it useful to construct an unbiased estimator of~$Y_H$ and take on some randomness in return for reduced computational cost.

Define the backward difference~$\Delta_n$ and represent the quantity of interest $Y_H$ with a telescoping series:
\begin{align*}
Y_H &= \sum_{n=1}^H \Delta_n & \text{where}\quad    \Delta_n &= \begin{cases}
    Y_n - Y_{n-1} & n > 1\\
    Y_1 & n = 1
    \end{cases}\,.
\end{align*}
We may sample from this telescoping series to provide unbiased estimates of~$Y_H$, introducing variance to our estimator in exchange for reducing expected computation.
We use the name \emph{randomized telescope} (RT) to refer to the family of estimators indexed by a distribution~$q$ over the integers~${1,\ldots,H}$ (for example, a geometric distribution) and a weight function $W(n, N)$:
\begin{align}\label{eq:rt_general}
\hat{Y}_H = \sum_{n=1}^N \Delta_n W(n, N) \quad \quad  N \in \{1,\ldots,H\} \sim q\,.
\end{align}
\begin{proposition}\label{prop:unbiased}
\textbf{Unbiasedness of RT estimators.}
The RT estimators in (\ref{eq:rt_general}) are unbiased estimators of
$Y_H$ as long as
\begin{align*}
\mathbb{E}_{N\sim q} [W(n, N) \mathds{1}\{N \geq n\}]\! =\!\! \sum_{N=n}^H\! W(n, N)q(N)\! =\! 1, \; \forall n\,.
\end{align*}
\end{proposition}
See Appendix B for a short proof.
Although we are coining the term ``randomized telescope'' to refer to the family of estimators with the form of Eq.~\ref{eq:rt_general}, the underlying trick has a long history, discussed in the next section.
The literature we are aware of focusses on one or both of two special cases of Eq.~\ref{eq:rt_general}, defined by choice of weight function $W(n, N)$.
We will also focus on these two variants of RT estimators, but we observe that there is a larger family.

Most related work uses the ``Russian roulette'' estimator originally discovered and named by von Neumann and Ulam \citep{kahn1955use}, which we term \emph{RT-RR} and has the form
\begin{align}
W(n, N) = \frac{1}{1 - \sum_{n'=1}^{n-1}q(n')} \mathds{1}\{N \geq n\}\,.
\end{align}
It can be seen as summing the iterates~$\Delta_n$ while flipping a biased coin at each iterate.
With probability~$q(n)$, the series is truncated at term~$N=n$.
With probability~${1 - q(n)}$, the process continues, and all future terms are upweighted by~$\frac{1}{1-q(n)}$ to maintain unbiasedness.

The other important special case of Eq.~\ref{eq:rt_general} is the ``single sample'' estimator \emph{RT-SS}, referred to as ``single term weighted truncation'' in \citet{lyne2015russian}.
RT-SS takes
\begin{align}
W(n, N) &= \frac{1}{q(N)} \mathds{1} \{n = N\}\,.
\end{align}
This is directly importance sampling the differences~$\Delta_n$.

We will later prove conditions under which RT-SS and RT-RR should be preferred.
Of all estimators in the form of Eq.~\ref{eq:rt_general} which obey proposition \ref{prop:unbiased} and for all~$q$, RT-SS minimizes the variance across worst-case diagonal covariances~$\Cov(\Delta_i, \Delta_j)$.
Within the same family, RT-RR achieves minimum variance when~$\Delta_i$ and~$\Delta_j$ are independent for all~$i, j$.

\subsection{A brief history of unbiased randomized truncation}
\label{sec:history}
The essential trick---unbiased estimation of a quantity via randomized truncation of a series---dates back to unpublished work from John von Neumann and Stanislaw Ulam.
They are credited for using it to develop a Monte Carlo method for matrix inversion in \citet{forsythe1950matrix}, and for a method for particle diffusion in \citet{kahn1955use}.

It has been applied and rediscovered in a number of fields and applications.
The early work from von Neumann and Ulam led to its use in computational physics, in neutron transport problems \citep{spanier1969monte}, for studying lattice fermions \citep{kuti1982stochastic}, and to estimate functional integrals \citep{wagner1987unbiased}.
In computer graphics \citet{arvo1990particle} introduced its use for ray tracing; it is now widely used in rendering software.
In statistical estimation, it has been used for estimation of derivatives \citep{rychlik1990unbiased}, unbiased kernel density estimation \citep{rychlik1995class}, doubly-intractable Bayesian posterior distributions \citep{girolami2013playing, lyne2015russian, wei2016markov}, and unbiased Markov chain Monte Carlo \citep{jacob2017unbiased}.

The underlying trick has been rediscovered by \citet{fearnhead2008particle} for unbiased estimation in particle filtering, by \citet{mcleish2010general} for debiasing Monte Carlo estimates, by \citet{rhee2012new, rhee2015unbiased} for unbiased estimation in stochastic differential equations,
and by \citet{tallec2017unbiasing} to debias truncated backpropagation.
The latter also uses RT estimators for optimization; however, it only considers fixed ``Russian roulette''-style randomized telescope estimators and does not consider convergence rates or how to adapt the estimator online (our main contributions).
\section{Optimizing loops and limits}
In this paper, we consider optimizing functions defined as limits.
Consider a problem where, given parameters~$\theta$ we can obtain a series of approximate losses~$\mathcal{L}_n(\theta)$, which converges uniformly to some limit~${\lim_{n \to H} \mathcal{L}_n := \mathcal{L}}$, for~${n\in\mathbb{N}_{+}}$ and~${H\in\mathbb{N}_{+}\cup\{\infty\}}$.
We assume the sequence of gradients with respect to~$\theta$, denoted ${G_n(\theta) := \nabla_\theta\mathcal{L}_n(\theta)}$ converge uniformly to a limit~${G(\theta)}$.
Under this uniform convergence and assuming convergence of~$\mathcal{L}_n$, we have~${\lim_{n \to H} \nabla_{\theta}\mathcal{L}_n(\theta) = \nabla_{\theta}\lim_{n \to H}\mathcal{L}_n(\theta)}$ (see Theorem 7.17 in \citet{rudin1976principles}), and so~${G(\theta)}$ is indeed the gradient of our objective~${\mathcal{L}(\theta)}$.
We assume there is a computational cost~$C(n)$ associated with evaluating~$\mathcal{L}_n$ or~$G_n$, nondecreasing with~$n$, and we wish to efficiently minimize~$\mathcal{L}$ with respect to~$\theta$.
Loops are an important special case of this framework, where~$\mathcal{L}_n$ is the final output resulting from running e.g., a training loop or RNN for some number of steps increasing in~$n$.

\subsection{Randomized telescopes for optimization}
We propose using randomized telescopes as a stochastic gradient estimator for such optimization problems.
We aim to accelerate optimization much as mini-batch stochastic gradient descent accelerates optimization for large datasets: using Monte Carlo sampling to decrease the expected cost of each optimization step, at the price of increasing variance in the gradient estimates, without introducing bias.

Consider the gradient~${G(\theta) = \lim_{n \to H} G_n(\theta)}$, and the backward difference ${\Delta_n(\theta) = G_n(\theta) - G_{n-1}(\theta)}$, where~${G_0(\theta) = 0}$, so that~${G(\theta) = \sum_{n=1}^H \Delta_n(\theta)}$.
We use the randomized telescope estimator
\begin{align}
\hat{G}(\theta) &= \sum_{n=1}^N\Delta_n(\theta) W(n,N)
\end{align}
where ${N \in \{1,2,\ldots,H\}}$ is drawn according to a proposal distribution~$q$, and together~$W$ and~$q$ satisfy proposition~\ref{prop:unbiased}.

Note that due to linearity of differentiation, and letting~${\mathcal{L}_0(\theta) := 0}$, we have
\begin{align*}
\sum_{n=1}^N\! \Delta_n(\theta) W(n, N) =
\nabla_\theta\!\! \sum_{n=1}^N\!
(\mathcal{L}_n(\theta)\! -\! \mathcal{L}_{n-1}(\theta))  W(n, N)\,.
\end{align*}
Thus, when the computation of $\mathcal{L}_n(\theta)$ can reuse most of the computation performed for~$\mathcal{L}_{n-1}(\theta)$, we can evaluate~$\hat{G}_N(\theta)$ via forward or backward automatic differentiation with cost approximately equal to computing~$G_N(\theta)$, i.e.,~$\hat{G}_N(\theta)$ has computation cost $\approx C(N)$.
This most often occurs when evaluating~$\mathcal{L}_n(\theta)$ involves an inner loop with a step size which does not change with $n$, e.g., meta-learning and training RNNs, but not solving ODEs or PDEs.
When computing~$\mathcal{L}_n(\theta)$ does not reuse computation evaluating~$\hat{G}_N(\theta)$ has computation cost~${\sum_{n=1}^N C(n) \mathds{1} \{W(n, N) \neq 0\}}$.

\subsection{Related work in optimization}
Gradient-based bilevel optimization has seen extensive work in literature.
See \citet{jameson1988aerodynamic} for an early example of optimizing implicit functions, \citet{christianson1998reverse} for a mathematical treatment, and \citet{maclaurin2015gradient, franceschi2017forward} for recent treatments in machine learning.
\citet{shaban2018truncated} propose truncating only the backward pass by only backpropagating through the final few optimization steps to reduce memory requirements.
\citet{metz2018learned} propose linearly increasing the number of inner steps over the course of the outer optimization.

An important case of bi-level optimization is optimization of architectures and hyperparameters.
Truncation causes bias, as shown by \citet{wu2018understanding} for learning rates and by \citet{metz2018learned} for neural optimizers.

Bi-level optimization is also used for meta-learning across related tasks \citep{schmidhuber1987evolutionary, bengio1992optimization}.
\citet{ravi2016optimization} train an initialization and optimizer, and \citet{finn2017model} only an initialization, to minimize validation loss.
The latter paper shows increasing performance with the number of steps used in the inner optimization.
However, in practice the number of inner loop steps must be kept small to allow training over many tasks.

Bi-level optimization can be accelerated by amortization.
Variational inference can be seen as bi-level optimization; variational autoencoders \citep{kingma2013auto} amortize the inner optimization with a predictive model of the solution to the inner objective.
Recent work such as \citet{brock2018smash, lorraine2018stochastic} amortizes hyperparameter optimization in a similar fashion.

However, amortizing the inner loop induces bias.
\citet{cremer2018inference} demonstrate this in VAEs, while \citet{kim2018semi} show that in VAEs, combining amortization with truncation by taking several gradient steps on the output of the encoder can reduce this bias.
This shows these techniques are orthogonal to our contributions: while fully amortizing the inner optimization causes bias, predictive models of the limit can accelerate convergence of $\mathcal{L}_n$ to $\mathcal{L}$.

Our work is also related to work on training sequence models.
\citet{tallec2017unbiasing} use the Russian roulette estimator to debias truncated backpropagation through time.
They use a fixed geometrically decaying~$q(N)$, and show that this improves validation loss for Penn Treebank.
They do not consider efficiency of optimization, or methods to automatically set or adapt the hyperparameters of the randomized telescope.
\citet{trinh2018learning} learn long term dependencies with auxiliary losses.
Other work accelerates optimization of sequence models by replacing recurrent models with models which use convolution or attention \citep{vaswani2017attention}, which can be trained more efficiently.
\section{Convergence rates with fixed RT estimators}
Before considering more complex large-scale problems, we examine the simple RT estimator for stochastic gradient descent on convex problems.
We assume that the sequence~$\mathcal{L}_n(\theta)$ and units for~$C$ are chosen such that~${C(n) = n}$.
We study RT-SS, with~$q(N)$ fixed \emph{a priori}.
We consider optimizing parameters~${\theta \in \mathcal{K}}$, where~${\mathcal{K} \subset \mathcal{R}^d}$ is a bounded, convex and compact set with diameter bounded by $D$.
We assume~$\mathcal{L}(\theta)$ is convex in~$\theta$, and~$G_n(\theta)$ converge according to~${||\Delta_n||_2 \leq \psi_n}$, where~$\psi_n$ converges polynomially or geometrically.
The quantity of interest is the instantaneous regret,~${R_t = \mathcal{L}(\theta_t) - \min_{\theta} \mathcal{L} (\theta)}$, where~$\theta_t$ is the parameter after~$t$ steps of SGD.

In this setting, any fixed truncation scheme using~$\mathcal{L}_N$ as a surrogate for~$\mathcal{L}$, with fixed ${N < H}$, cannot achieve~${\lim_{t \to \infty} R_t = 0}$.
Meanwhile, the fully unrolled estimator has computational cost which scales with~$H$.
In the many situations where~${H \to \infty}$, it is impossible to take even a single gradient step with this estimator.

The randomized telescope estimator overcomes these drawbacks by exploiting the fact that~$G_n$ converges according to~$||\Delta_n||_2 \leq \psi_n$.
As long as~$q$ is chosen to have tails no lighter than~$\psi_n$, for sufficiently fast convergence, the resulting RT-SS gradient estimator achieves asymptotic regret bounds invariant to~$H$ in terms of convergence rate.

All proofs are deferred to Appendix B.
We begin by proving bounds on the variance and expected computation for polynomially decaying~$q(N)$ and~$\psi_n$.
\begin{theorem}\label{thm:poly}
\textbf{Bounded variance and compute with polynomial convergence of $\psi$}.
Assume~$\psi$ converges according to~${\psi_n \leq \nicefrac{c}{n^p}}$ or faster, for constants~${p > 0}$ and~${c > 0}$.
Choose the RT-SS estimator with~${q(N) \propto 1/(N^{p + 1/2})}$.
The resulting estimator~$\hat{G}$ achieves expected compute~${C \leq (\mathcal{H}_{H}^{p-\nicefrac{1}{2}})^2}$, where~$\mathcal{H}_H^i$ is the~$H$th generalized harmonic number of order~$i$, and expected squared norm~${\mathbb{E}[ ||\hat{G}||_2^2 ] \leq c_{\psi}^2 (\mathcal{H}_H^{p-\nicefrac{1}{2}})^2 := \tilde{G}^2}$.
The limit~${\lim_{H \to \infty} \mathcal{H}_H^{p - \nicefrac{1}{2}}}$ is finite iff~${p > \nicefrac{3}{2}}$, in which case it is given by the Riemannian zeta function,~${\lim_{H \to \infty} \mathcal{H}_H^{p - \nicefrac{1}{2}} = \zeta(p - \nicefrac{1}{2})}$.
Accordingly, the estimator achieves horizon-agnostic variance and expected compute bounds iff~${p > \nicefrac{3}{2}}$.
\end{theorem}

The corresponding bounds for geometrically decaying $q(N)$ and $\psi_n$ follow.

\begin{theorem}\label{thm:geom}
\textbf{Bounded variance and compute with geometric convergence of $\psi$}.
Assume~$\psi_n$ converges according to~${\psi_n \leq c p^n}$, or faster, for~${0 < p < 1}$.
Choose RT-SS and with~${q(N) \propto p^N}$.
The resulting estimator ~$\hat{G}$ achieves expected compute~${C \leq (1-p)^{-2}}$ and expected squared norm~${||\hat{G}||_2^2 \leq \frac{c}{(1-p)^2} := \tilde{G}^2}$.
Thus, the estimator achieves horizon-agnostic variance and expected compute bounds for all~${0 < p < 1}$.
\end{theorem}
Given a setting and estimator~$\hat{G}$ from either \ref{thm:poly} or \ref{thm:geom}, with corresponding expected compute cost~$C$ and upper bound on expected squared norm~$\tilde{G}^2$, the following theorem considers regret guarantees when using this estimator to perform stochastic gradient descent.

\begin{theorem}\label{thm:infopt}
\textbf{Asymptotic regret bounds for optimizing infinite-horizon programs}.
Assume the setting from \ref{thm:poly} or \ref{thm:geom}, and the corresponding~$C$ and~$\tilde{G}$ from those theorems.
Let~$R_t$ be the instantaneous regret at the~$t$th step of optimization,~${R_t = \mathcal{L}(\theta_t) - \min_\theta \mathcal{L} (\theta)}$.
Let~$t(B)$ be the greatest~$t$ such that a computational budget~$B$ is not exceeded.
Use online gradient descent with step size~${\eta_t = \frac{D}{\sqrt{t} \mathbb{E}[||\hat{G}||_2^2]}}$.
As~${B \to \infty}$, the asymptotic instantaneous regret is bounded by~${R_{t(B)} \leq \mathcal{O} (\tilde{G} D \sqrt{\frac{C}{B}})}$, independent of~$H$.
\end{theorem}

Theorem \ref{thm:infopt} indicates that if~$G_n$ converges sufficiently fast and~$\mathcal{L}_n$ is convex, the RT estimator provably optimizes the limit.

\section{Adaptive RT estimators}
In practice, the estimator considered in the previous section may have high variance.
This section develops an objective for designing such estimators, and derives closed-form~$W(n, N)$ and~$q$ which maximize this objective given estimates of~$\mathbb{E}[||\Delta_i||_2^2]$ and assumptions on~$\Cov(\Delta_i, \Delta_j)$.

\subsection{Choosing between unbiased estimators}
We propose choosing an estimator which achieves the best lower bound on the expected improvement per compute unit spent, given smoothness assumptions on the loss.
Our analysis builds on that of \citet{balles2016coupling}: they adaptively choose a batch size using batch covariance information, while we choose between between arbitrary unbiased gradient estimators using knowledge of those estimators' expected squared norms and computation cost.

Here we assume that the true gradient of the objective~${\nabla_\theta [ \mathcal{L}(\theta)] := \nabla_\theta}$ (for compactness of notation) is smooth in~$\theta$.
We do not assume convexity.
Note that ~$\nabla_\theta$ is not necessarily equal to~$G(\theta)$, as the loss~$\mathcal{L}(\theta)$ and its gradient~$G(\theta)$ may be random variables due to sampling of data and/or latent variables.

We assume that~$\mathcal{L}$ is~$L$-smooth (the gradients of~$\mathcal{L}(\theta)$ are $L$-Lipschitz), i.e., there exists a constant~${L > 0}$ such that ${\nabla_{\theta_b}\! -\! \nabla_{\theta_a}\! \leq\! L || \theta_b\! -\! \theta_a||_2 \quad \forall \theta_a, \theta_b \in \mathbb{R}^d}$.
It follows \citep{balles2016coupling, bottou2018optimization} that, when performing SGD with an unbiased stochastic gradient estimator~$\hat{G}_t$,
\begin{multline}
\mathbb{E}[\mathcal{L}_H(\theta_t) - \mathcal{L}_H (\theta_{t+1})] \\ \geq
\mathbb{E}[\eta_t \nabla_{\theta_t}^T \hat{G}_t(\theta_{t})] - \mathbb{E}[\frac{L\eta_t^2}{2} ||\hat{G}_t(\theta_{t})||_2^2]\,.
\end{multline}
Unbiasedness of $\hat{G}$ implies
$\mathbb{E}[\nabla_{\theta_t}^T \hat{G}_t(\theta_t)] = ||\nabla_{\theta_t}^T||_2^2$,
thus:
\begin{multline}
\mathbb{E}[\mathcal{L}_H(\theta_t) - \mathcal{L}_H (\theta_{t+1})]
\\ \geq
\mathbb{E}[\eta_t ||\nabla_\theta||_2^2] - \mathbb{E}[\frac{L\eta_t^2}{2} ||\hat{G}_t(\theta_{t})||_2^2]
:= J\,.
\end{multline}
Above, $J$ is a lower bound on the expected improvement in the loss from one optimization step.
Given a fixed choice of~$\hat{G}_t(\theta_t)$, how should one pick the learning rate~$\eta_t$ to maximize~$J$ and what is the corresponding lower bound on expected improvement?

Optimizing~$\eta_t$ by finding~$\eta_t^\star$ s.t.~${\nicefrac{dJ}{d\eta_t^\star} = 0}$ yields
\begin{align}
\eta_t^\star &= \frac{||\nabla_\theta||_2^2}{L \mathbb{E}[||\hat{G}_t(\theta_{t})||_2^2]} \propto \frac{1}{\mathbb{E}[||\hat{G}_t(\theta_{t})||_2^2]}\\
J^\star &= \frac{||\nabla_\theta||^4}{2 L \mathbb{E}[||\hat{G}_t(\theta_{t})||_2^2]} \propto \frac{1}{\mathbb{E}[||\hat{G}_t(\theta_{t})||_2^2]}\,.
\end{align}
This tells us how to choose~$\eta_t$ if we know~$L$,~$||\hat{G}||_2^2$, etc.
In practice, it is unlikely that we know~$L$ or even~$||\nabla_{\theta_t}||_2$.
We instead assume we have access to some ``reference'' learning rate~$\bar{\eta_t}$, which has been optimized for use with a ``reference'' gradient estimator~$\bar{G}_t$, with known~$\mathbb{E}[||\bar{G}_t||_2^2]$.
When using RT estimators, we may have access to learning rates which have been optimized for use with the un-truncated estimator.
Even when we do not know an optimal reference learning rate, this construction means we need only tune one hyperparameter (the reference learning rate), and can still choose between a family of gradient estimators online.
Instead of directly maximizing~$J$, we choose~$\eta_t$ for~$\hat{G}$ by maximizing \textit{improvement relative to the reference estimator} in terms of~$J$, the lower bound on expected improvement.

Assume that~$\bar{\eta}_t$ has been set optimally for a problem and reference estimator~$\bar{G}$ up to some constant~$k$, i.e.,
\begin{align}
\bar{\eta}_t &= k \frac{||\nabla_{\theta_t}||_2^2}{L \mathbb{E}[||\bar{G}_t(\theta_{t})||_2^2]}\,.
\end{align}
\vspace{-1.25\baselineskip}

Then the expected improvement~$\bar{J}$ obtained by the reference estimator~$\bar{G}$ is:
\vspace{-1.75\baselineskip}

\begin{align}
\bar{J} &= (k - \frac{k^2}{2}) \frac{||\nabla_{\theta_t}||^4}{2 L \mathbb{E}[||\bar{G}_t(\theta_{t})||_2^2]}
\end{align}
\vspace{-1.25\baselineskip}

We assume that~${0 < k < 2}$, such that~$\bar{J}$ is positive and the reference has guaranteed expected improvement.
Now set the learning rate according to
\vspace{-1.75\baselineskip}

\begin{align}
\label{eq:rel-lr}
\eta_t &=
\bar{\eta_t} \frac{\mathbb{E}[||\hat{G}_t||_2^2]}{\mathbb{E}[||\bar{G}_t||_2^2]}\,.
\end{align}
\vspace{-1.25\baselineskip}

It follows that the expected improvement~$\hat{J}$ obtained by the estimator~$\hat{G}$ is
\vspace{-1.75\baselineskip}

\begin{align}
\hat{J} &= \frac{\mathbb{E}[||\bar{G}_t(\theta_{t})||_2^2]}{\mathbb{E}[||\hat{G}_t(\theta_{t})||_2^2]} \bar{J}
\end{align}
\vspace{-1.25\baselineskip}

Let the expected computation cost of evaluating~$\hat{G}$ be~$\hat{C}$.
We want to maximize~$\nicefrac{\hat{J}}{\hat{C}}$.
If we use the above method to choose~$\eta_t$, we have~${\nicefrac{\hat{J}}{\hat{C}} \propto \big(\hat{C} \mathbb{E} ||\hat{G}_t(\theta_{t})||_2^2 \big)^{-1}}$.
We call~${\big(\hat{C} \mathbb{E} ||\hat{G}_t(\theta_{t})||_2^2\big)^{-1}}$ the \textit{relative optimization efficiency}, or ROE.
We decide between gradient estimators~$\hat{G}$ by choosing the one which maximizes the ROE.
Once an estimator is chosen, one should choose a learning rate according to (\ref{eq:rel-lr}) relative to a reference learning rate~$\bar{\eta}$ and estimator~$\bar{G}$.

\subsection{Optimal weighted sampling for RT estimators}
Now that we have an objective, we can consider designing RT estimators which optimize the ROE.
For the classes of single sample and Russian roulette estimators, we prove conditions under which that class maximizes the ROE across an arbitrary choice of RT estimators.
We also derive closed-form expressions for the optimal sampling distribution~$q$ for each class, under the conditions where that class is optimal.

We assume that computation can be reused and evaluating~${\hat{G}_H = \sum_{n=1}^N \Delta_n W(n, N)}$ has computation cost~$C(N)$.
As described in Section 3.1, this is approximately true for many objectives.
When it is not, the cost of computing~$\sum_{n=1}^N \Delta_n W(n, N)$ is~${\sum_{n=1}^N C(n) \mathds{1}\{(W(n, N) \neq 0) \text{ or } (W(n+1, N) \neq 0) \}}$.
This would penalize the ROE of dense ~$W(n, N)$ and favor sparse~$W(n, N)$, possibly impacting the optimality conditions for RT-RR.
We mitigate this inaccuracy by subsequence selection (described in the following subsection), which allows construction of sparse sampling strategies.

We begin by showing the RT-SS estimator is optimal with regards to worst-case diagonal covariances $\Cov(\Delta_i, \Delta_j)$, and deriving the optimal $q(N)$.
\begin{theorem}\label{thm:advcorr-w} \textbf{Optimality of RT-SS under adversarial correlation.}
Consider the family of estimators presented in Equation  \ref{eq:rt_general}.
Assume $\theta$, $\nabla_\theta$, and $G$ are univariate.
For any fixed sampling distribution $q$, the single-sample RT estimator RT-SS minimizes the worst-case variance of $\hat{G}$ across an adversarial choice of covariances~${\Cov(\Delta_i, \Delta_j) \leq \sqrt{\Var(\Delta_i)} \sqrt{\Var(\Delta_j)}}$.
\end{theorem}

\begin{theorem}\label{thm:advcorr-q} \textbf{Optimal q under adversarial correlation.}
Consider the family of estimators presented in Equation  \ref{eq:rt_general}.
Assume $\Cov(\Delta_i, \Delta_i)$ and $\Cov(\Delta_i, \Delta_j)$ are diagonal.
The RT-SS estimator with~${q_n \propto \sqrt{\frac{\mathbb{E}[||\Delta_n||_2^2}{C(n)}}}$ maximizes the ROE across an adversarial choice of diagonal covariance matrices~${\Cov(\Delta_i, \Delta_j)_{kk} \leq \sqrt{\Cov(\Delta_i, \Delta_i)_{kk} \Cov(\Delta_j, \Delta_j)_{kk}}}$.
\end{theorem}
We next show the RT-RR estimator is optimal when~$\Cov(\Delta_i, \Delta_i)$ is diagonal and~$\Delta_i$ and~$\Delta_j$ are independent for~${j \neq i}$, and derive the optimal~$q(N)$.

\begin{theorem}\label{thm:nocorr-w} \textbf{Optimality of RT-RR under independence}.
Consider the family of estimators presented in Eq.~\ref{eq:rt_general}.
Assume the $\Delta_j$ are univariate.
When the~$\Delta_j$ are uncorrelated, for any importance sampling distribution~$q$, the Russian roulette estimator achieves the minimum variance in this family and thus maximizes the optimization efficiency lower bound.
\end{theorem}
\begin{theorem}\label{thm:nocorr-q} \textbf{Optimal q under independence}.
Consider the family of estimators presented in Equation  \ref{eq:rt_general}.
Assume~$\Cov(\Delta_i, \Delta_i)$ is diagonal and~$\Delta_i$ and~$\Delta_j$ are independent.
The RT-RR estimator with~${Q(i) \propto \sqrt{\frac{\mathbb{E} [||\Delta_i||_2^2}{C(i) - C(i-1)}]}}$, where~${Q(i) = \Pr(n \geq i) = \sum_{j=i}^H q(j)}$,
maximizes the ROE.
\end{theorem}

\subsection{Subsequence selection}
The scheme for designing RT estimators given in the previous subsection
contains assumptions which will often not hold in practice. To partially
alleviate these concerns, we can design the \textit{sequence of iterates over
which we apply the RT estimator} to maximize the ROE.

Some sequences may result in more efficient estimators, depending on how the intermediate iterates $G_n$ correlate with $G$.
The variance of the estimator, and the ROE, will be reduced if we choose a sequence $\mathcal{L}_n$ such that $G_n$ is positively correlated with $G$ for all $n$.

We begin with a reference sequence~$\bar{\mathcal{L}}_i$, $\bar{G}_i$, with cost function~$\bar{C}$, where~${i, j \in \mathcal{N}}$ and~${i, j \leq \bar{H}}$,
and where~$\bar{G}_i$ has cost~$\bar{c}_i$.
We assume knowledge of~${\mathbb{E}[||\bar{G}_i \!-\! \bar{G}_j||_2^2]}$.
We aim to find a subsequence~${S \in \mathcal{S}}$, where~$\mathcal{S}$ is the
set of subsequences over the integers~$1,\ldots,\bar{H}$ which have final element~${S_{-1} = \bar{H}}$.
Given~$S$, we take~${\mathcal{L}_n = \bar{\mathcal{L}}_{S_n}}$, ${G_n = \bar{G}_{S_n}}$, ${C(n) = \bar{C}(S_n)}$, ${H = |S|}$, and~${\Delta_n = G_n - G_{n-1}}$, where~${G_0 := 0}$.

In practice, we greedily construct~$S$ by adding
indexes~$i$ to the sequence~$[\bar{H}]$ or removing indexes~$i$ from the
sequence~$[1, \ldots, \bar{H}]$. As this step requires minimal computation, we
perform both greedy adding and greedy removal and return the~$S$ with the best
ROE. The minimal subsequence~${S = [\bar{H}]}$ is always considered,
allowing RT estimators to fall back on the original full-horizon estimator.
\section{Practical implementation}
\subsection{Tuning the estimator}
We estimate the expected squared distances
$\mathbb{E}[||\bar{G}_i - \bar{G}_j||_2^2]$ by
maintaining exponential moving averages. We keep track of the computational
budget $B$ used so far by the RT estimator, and ``tune'' the estimator
every $K \bar{C}(\bar{H})$ units of computation, where
$\bar{C}(\bar{H})$ is the compute required to
evaluate $\bar{G}_{\bar{H}}$, and $K$ is a ``tuning frequency'' hyperparameter.
During tuning, the gradients $G_i$ are computed, the squared norms
$||\bar{G}_i - \bar{G}_j||_2^2$ are computed, and the exponential moving averages
are updated. At the end of tuning, the estimator is updated using the
expected squared norms; i.e. a subsequence is selected, $q$ is set according
to section 5.2 with choice of RT-RR or RT-SS left as a hyperparameter, and
the learning rate is adapted according to section 5.1

\subsection{Controlling sequence length}
Tuning and subsequence selection require computation. Consider using RT
to optimize an objective with an inner loop of size $M$.
If we let $\bar{G}_i$
be the gradient of the loss after $i$ inner steps,
we must maintain $M^2 - M$ exponential moving
averages $\mathbb{E}||\bar{G}_i - \bar{G}_j||_2^2$, and
compute $M$ gradients $\bar{G}_i$ each time we tune the estimator.
The computational cost of the tuning step under this
scheme is $\mathcal{O}(M^2)$.
This is unacceptable if we wish our method to scale well
with the size of loops we might wish to optimize.

To circumvent this, we choose base subsequences such that~${\bar{C}_i \propto 2^i}$. This ensures that~${\bar{H} = \mathcal{O}(\log_2 M)}$, where~$M$ is the maximum number of steps we
wish to unroll. We must maintain~$\mathcal{O} (\log_2^2 M)$ exponential moving
averages. Computing the gradients~$\bar{G}_i$ during each tuning step
requires compute~${C_{\text{tune}} = \sum_{i=1}^{\bar{H}} k * 2^i}$. Noting
that~${\bar{C}_{\bar{H}} = k * 2^{\bar{H}}}$ and that~${\sum_{i=1}^N 2^i < 2^{N+1} \forall N}$ yields~${C_{\text{tune}} < 2 \bar{C}_{\bar{H}} = 2 M}$.

\section{Experiments}
For all experiments, we tune learning rates for the full-horizon un-truncated estimator via grid search over all~${a \times 10^{-b}}$, for~${a \in \{1.0, 2.2, 5.5\}}$ and~${b \in \{0.0, 1.0, 2.0, 3.0, 5.0\}}$.
The same learning rates are used for the truncated estimators and (as reference learning rates) for the RT estimators.
We do not decay the learning rate. Experiments are run with the random seeds $0, 1, 2, 3, 4$ and we plot means and standard deviation.

We use the same hyperparameters for our online tuning procedure for all experiments: the tuning frequency $K$ is set to $5$, and the exponential moving average weight $\alpha$ is set to $0.9$.
These hyperparameters were not extensively tuned.
For each problem, we compare deterministic, RT-SS, and RT-RR estimators, each with a range of truncations.

\subsection{Lotka-Volterra ODE}
We first experiment with variational inference of parameters of a Lotka-Volterra (LV) ODE. LV ODEs are defined by the predator-prey equations, where~$u_2$ and~$u_1$ are predator and prey populations, respectively:
\vspace{-1.5\baselineskip}

\begin{align*}
\frac{du_1}{dt} &= A u_1 - B u_1 u_2 &
\frac{du_2}{dt} &= C u_1 u_2 - D u_2
\end{align*}
\vspace{-1.5\baselineskip}

We aim to infer the parameters $\lambda = [u_1(t=0), u_2(t=0), A, B, C, D]$.
The true parameters are drawn from $\mathcal{U}([1.0, 0.4, 0.8, 0.4, 1.5, 0.4], [1.5, 0.6, 1.2, 0.6, 2.0, 0.6])$, chosen empirically to ensure stability solving the equations.
We generate ground-truth data by solving the equations using RK4 (a common 4th-order Runge Kutta method) from~${t = 0}$ to~${t = 5}$ with~$10000$ steps.
The learner is given access to five equally spaced noisy observations~$y(t)$, generated according to~${y(t) = u(t) + \mathcal{N}(0, 0.1)}$.

We place a diagonal Gaussian prior on $\theta$ with the same mean and standard deviation as the data-generating distribution.
The variational posterior is a diagonal Gaussian~$q(\lambda)$ with mean~$\mu$ and standard deviation~$\sigma$.
The parameters optimized are ${\theta = [\tilde{\mu}, \tilde{\sigma}]}$.
We let~${\mu = g(\tilde{\mu})}$ and~${\sigma = g(\tilde{\sigma})}$, where~${g(\tilde{x}) = \log(1+e^{\tilde{x}})}$, to ensure positivity.
We use a reflecting boundary to ensure positivity of parameter samples from $q$.
The variational posterior is initialized to have mean equal to the prior and standard deviation 0.1.

The loss considered is the negative evidence lower bound (negative ELBO). The ELBO is:
\begin{align*}
\text{ELBO}(q(\theta))\!&=\! \mathbb{E}_{q(\theta)}\! \sum_{t}\! \log p(y(t) | u_\theta(t))\! +\! D_{\sf{KL}}\big(q(\theta) || p(\theta) \big)
\end{align*}
\vspace{-1.5\baselineskip}

Above, $u_\theta(t)$ is the value of the solution $u_\theta$ to the LV ODE with parameters $\theta$, evaluated at time $t$.
We consider a sequence $\mathcal{L}_n(\theta)$, where in computing the ELBO, $u_\theta(t)$ is approximated by solving the ODE using RK4 with ${2^n + 1}$ steps,
and linearly interpolating the solution to the $5$ observation times.
The outer-loop optimization is performed with a batch size of $64$ (i.e., $64$ samples of $\theta$ are performed at each step) and a learning rate of $0.01$.
Evaluation is performed with a batch size of $512$.

\begin{figure}
\vspace{-0.3cm}
\small
\begin{tabular}{c c}
\rotatebox{90}{\qquad\qquad$\text{Negative ELBO}$}&
\hspace{-2mm}\includegraphics[width=0.9\linewidth, clip, trim=2mm 2mm 0cm 0cm]{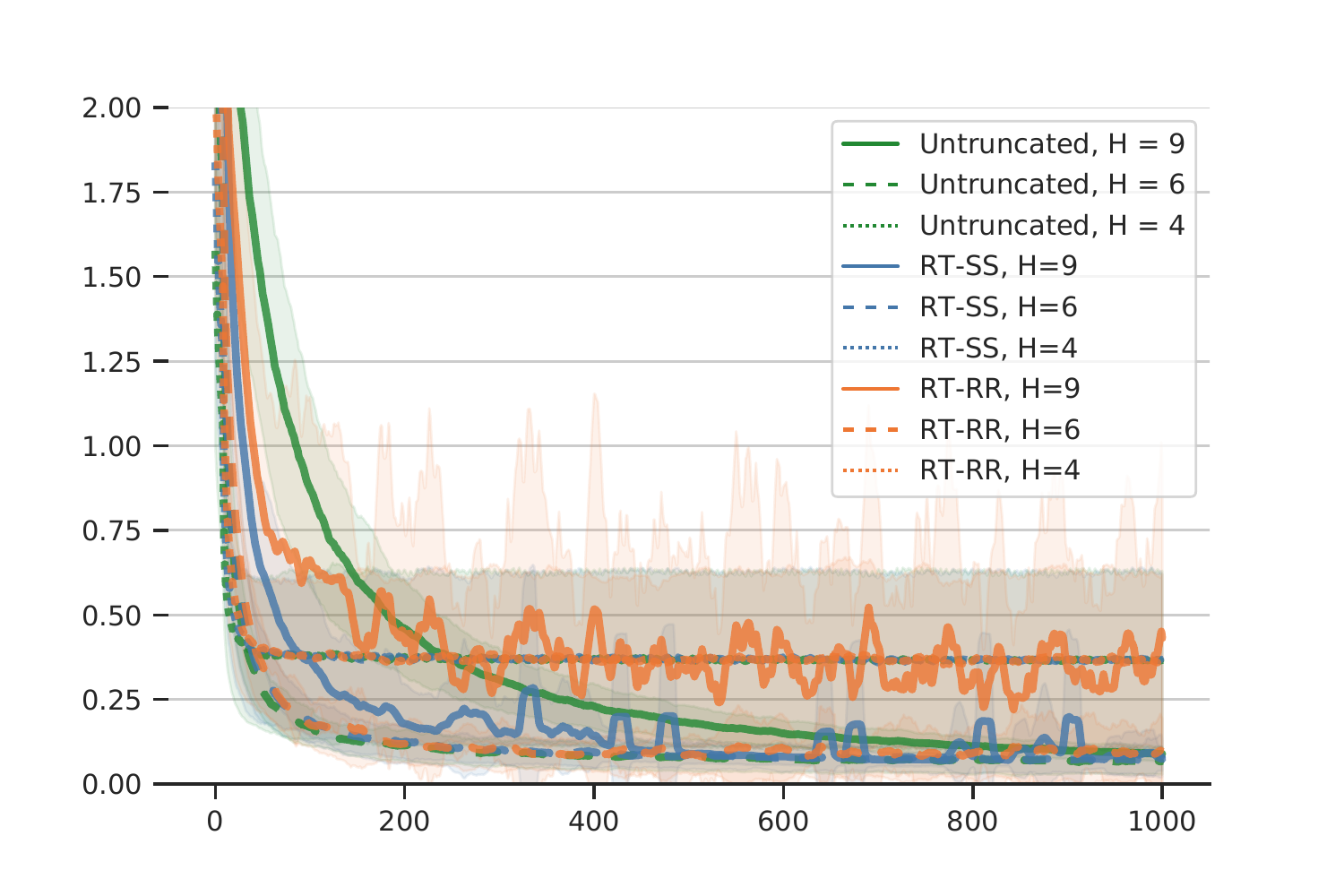} \vspace{-1mm} \\
& \vspace{-2mm} Gradient evaluations (thousands)
\label{fig:lv_loss}
\end{tabular}
\caption{Lotka-Volterra parameter inference}
\vspace{-0.5cm}
\end{figure}

Figure \ref{fig:lv_loss} shows the loss of the different estimators over the course of training.
RT-SS estimators outperform the un-truncated estimator without inducing bias.
They are competitive with the truncation~${H = 6}$, while avoiding the bias present with the truncation~${H = 4}$, at the cost of some variance.
Some RT-RR estimators experience issues with optimization, appearing to obtain the same biased solution as the ${H = 4}$ truncation.

\subsection{MNIST learning rate}
We next experiment with meta-optimization of a learning rate on MNIST.
We largely follow the procedure used by \citet{wu2018understanding}.
We use a feedforward network with two hidden layers of 100 units, with weights initialized from a Gaussian with standard deviation 0.1, and biases initialized to zero.
Optimization is performed with a batch size of 100.

The neural network is trained by SGD with momentum using Polyak averaging, with the momentum parameter fixed to~$0.9$.
We aim to learn a learning rate~$\eta_0$ and decay~$\lambda$ for the inner-loop optimization.
These are initialized to~$0.01$ and~$0.1$ respectively.
The learning rate for the inner optimization at an inner optimization step~$t$ is~$
{\eta_t = \eta_0(1 + \frac{t}{5000})^{-\lambda}}$.

As in \citet{wu2018understanding}, we pre-train the net for 50 steps with a learning rate of~$0.1$.
$\mathcal{L}_n$ is the evaluation loss after~${2^n + 1}$ training steps with a batch size of~$100$.
The evaluation loss is measured over~${2^n + 1}$ validation batches or the entire validation set, whichever is smaller.
The outer optimization is performed with a learning rate of~$0.01$.

RT-SS estimators achieve faster convergence than fixed-truncation estimators.
RT-RR estimators suffer from very poor convergence.
Truncated estimators appear to obtain biased solutions.
The un-truncated estimator achieves a slightly better loss than the RT estimators, but takes significantly longer to converge.

\begin{figure}
\vspace{-0.3cm}
\small
\begin{tabular}{c c}
\rotatebox{90}{\qquad\qquad$\text{Evaluation loss}$}&
\hspace{-2mm}\includegraphics[width=0.9\linewidth, clip, trim=2mm 2mm 0cm 0cm]{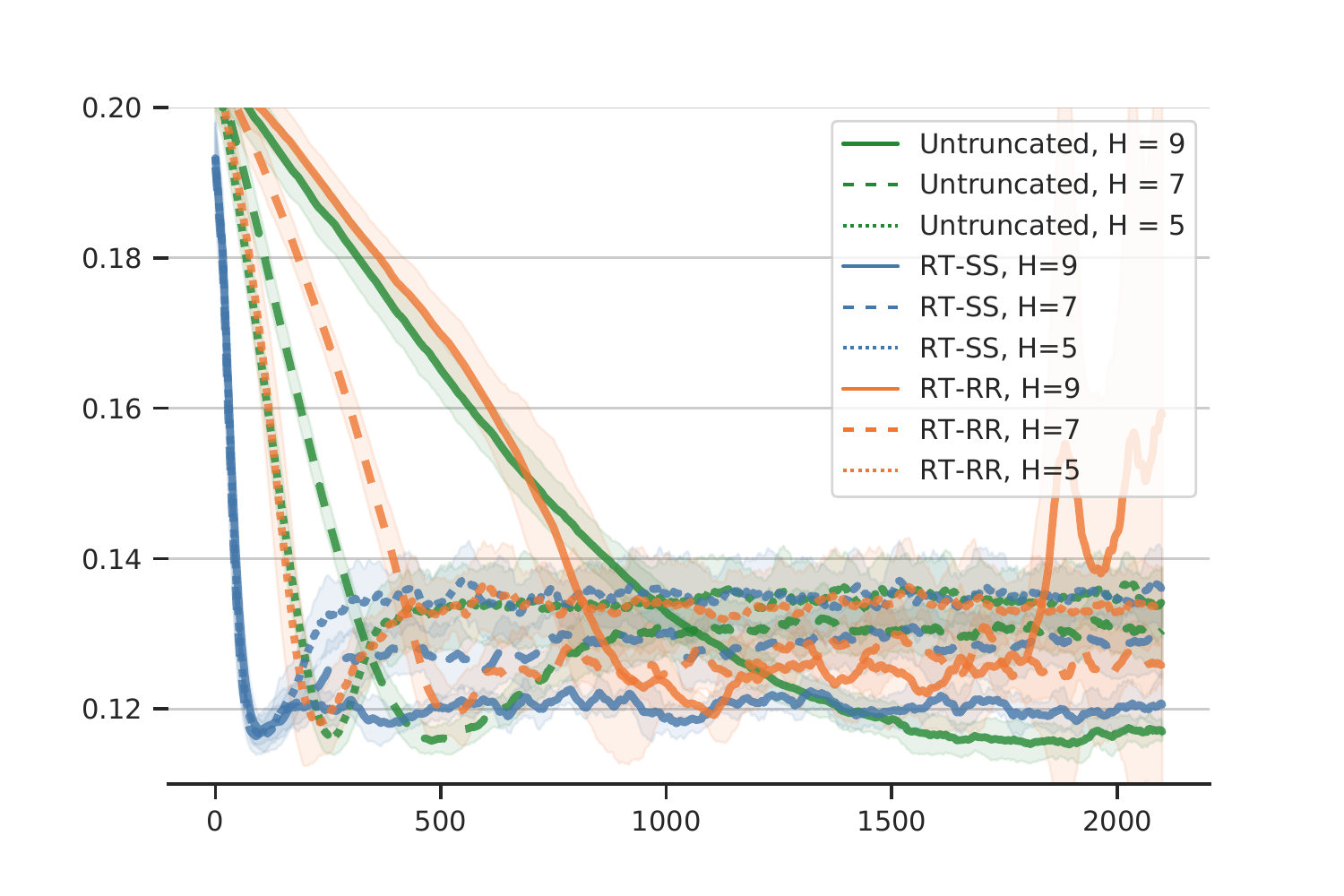} \vspace{-1mm} \\
& \vspace{-2mm} Neural network evaluations (thousands)
\label{fig:mnist_loss}
\end{tabular}
\caption{MNIST learning rate meta-optimization}
\vspace{-0.75cm}
\end{figure}

\subsection{\texttt{enwik8} LSTM}
Finally, we study a high-dimensional optimization problem:
training an LSTM to model sequences on $\texttt{enwik8}$.
These data are the first 100M bytes of a Wikipedia XML dump.
There are 205 unique tokens. We use the first 90M,
5M, and 5M characters as the training, evaluation, and test sets.

We build on code\footnote{\url{http://github.com/salesforce/awd-lstm-lm}} from
\citet{merity2017regularizing, merity2018analysis}.
We train
an LSTM with 1000 hidden units and 400-dimensional input and output
embeddings. The model has 5.9M parameters. The only regularization is
an $\ell_2$ penalty on the weights with magnitude $10^{-6}$.
The optimization is performed with a learning rate of $2.2$.
This model is not state-of-the-art: our aim to investigate performance of
RT estimators for optimizing high-dimensional neural networks, rather than to
maximize performance at a language modeling task.

We choose $\mathcal{L}_n$ to be the mean
cross-entropy after unrolling the LSTM training for ${2^{n-1} + 1}$ steps.
We choose the horizon ${H = 9}$, such that
the un-truncated loop has 257 steps, chosen to be close to the
200-length training sequences used by \citet{merity2018analysis}.

\begin{figure}
\vspace{-0.3cm}
\small
\begin{tabular}{c c}
\rotatebox{90}{\qquad$\text{Training bits-per-character}$}&
\hspace{-2mm}\includegraphics[width=0.9\linewidth, clip, trim=2mm 2mm 0cm 0cm]
{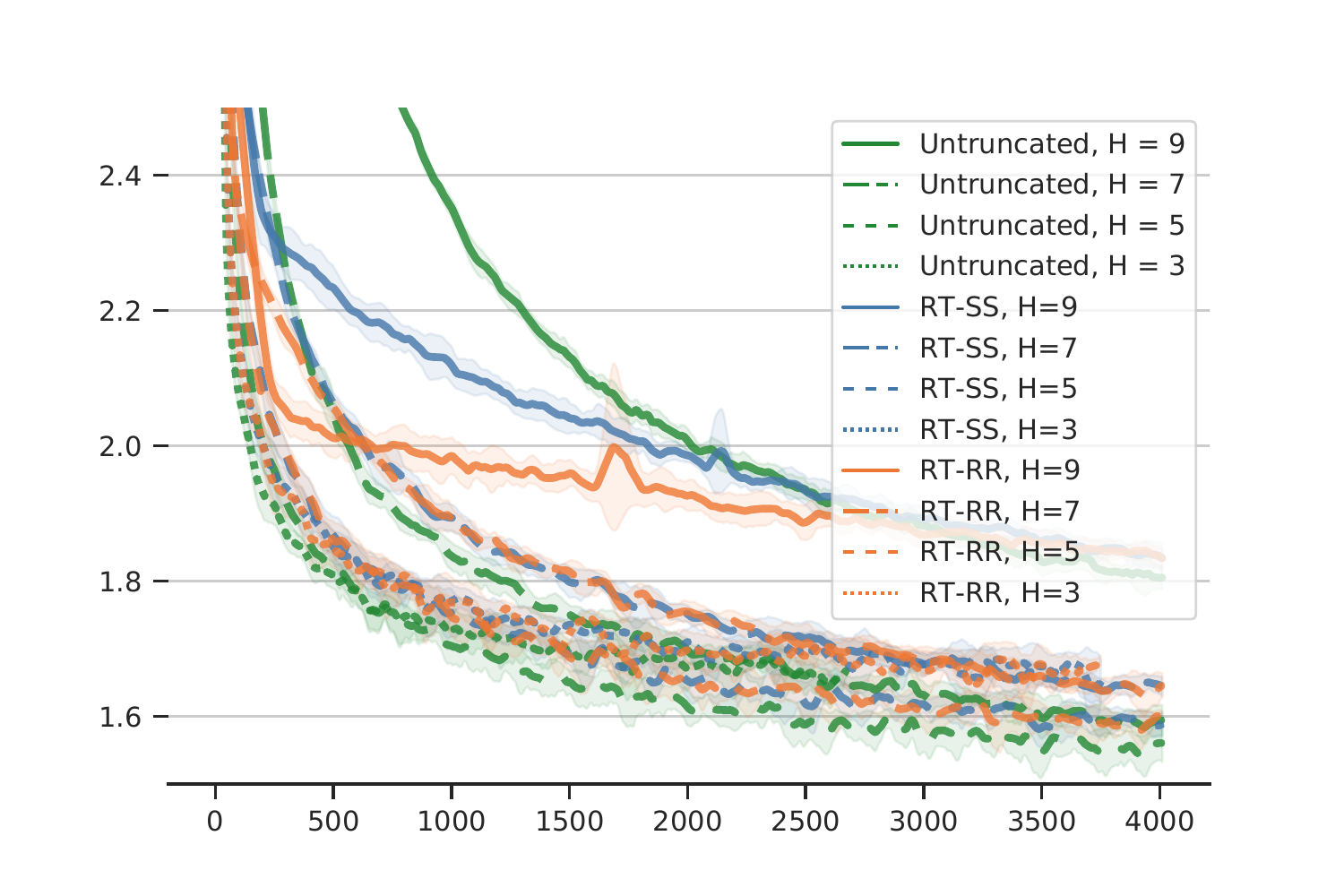} \vspace{-1mm} \\
& \vspace{-2mm} LSTM cell evaluations (thousands)
\label{fig:enwik}
\end{tabular}
\caption{LSTM training on \texttt{enwik8}}
\vspace{-0.9cm}
\end{figure}

Figure \ref{fig:enwik} shows the training bits-per-character (proportional to the training cross-entropy loss).
RT estimators provide some acceleration over the un-truncated~${H=9}$ estimator early in training, but after about 200k cell evaluations, fall back on the un-truncated estimator,
subsequently progressing slightly more slowly due to computational cost of tuning.
We conjecture that the diagonal covariance assumption in Section 5 is unsuited to high-dimensional problems, and leads to overly conservative estimators.

\section{Limitations and future work}
\textbf{Other optimizers}.
We develop the lower bound on expected improvement for SGD.
Important future directions would investigate adaptive and momentum-based SGD methods such as Adam \citep{kingma2014adam}.

\textbf{Tuning step}.
Our method includes a tuning step which requires computation.
It might be possible to remove this tuning step by estimating covariance structure online using just the values of $\hat{G}$ observed during each optimization step.

\textbf{RT estimators beyond RT-SS and RT-RR}.
There is a rich family defined by choices of $q$ and $W(n, N)$.
The optimal member depends on covariance structure between the $G_i$.
We explore RT-SS and RT-RR under strict covariance assumptions.
Relaxing these assumptions and optimizing $q$ and $W$ across a wider family could improve adaptive estimator performance for high-dimensional problems such as training RNNs.

\textbf{Predictive models of the sequence limit}.
Using any sequence $G_n$ with RT yields an unbiased estimator as long as the sequence is consistent, i.e. its limit $G$ is the true gradient.
Combining randomized telescopes with predictive models of the gradients \citep{jaderberg2017decoupled, weber2019credit} might yield a fast-converging sequence, leading to estimators with low computation and variance.

\section{Conclusion}
We investigated the use of randomly truncated unbiased gradient estimators for
optimizing objectives which involve loops and limits.
We proved these estimators can achieve
horizon-independent convergence rates for optimizing loops and limits.
We derived adaptive variants which can be tuned online
to maximize a lower bound on expected improvement per unit
computation. Experimental results matched theoretical intuitions that the
single sample estimator is more robust than Russian roulette for optimization.
The adaptive RT-SS estimator often significantly accelerates
optimization, and can otherwise fall back on the un-truncated estimator.

\section{Acknowledgements}
We would like to thank Matthew Johnson, Peter Orbanz, and James Saunderson for helpful discussions.
This work was funded by the Alfred P.\ Sloan Foundation and NSF IIS-1421780.

\bibliography{references}
\bibliographystyle{icml2019}

\appendix
\onecolumn
\clearpage
\section{Algorithm pseudocode}

\begin{algorithm}[H]
   \caption{Optimization with randomized telescopes}
   \label{alg:opt}
\begin{algorithmic}
   \STATE {\bfseries Input:} initial parameter $\theta$,
   gradient routine $g(\theta, i)$ which returns $\bar{G}_i(\theta)$,
   compute costs $\bar{C}$,
   exponential decay $\alpha$, tuning frequency $K$, horizon $\bar{H}$,
   reference learning rate $\bar{\eta}$
   \STATE Initialize $B = 0$, next\_tune$ = 0$, $D_{i, j} = 0$
   \REPEAT
   \IF{next\_tune$ <= B$}
   \STATE $\bar{D}, q, W, S \leftarrow \text{tune}(
   \theta, \bar{D}, g, \bar{C}, \alpha, \bar{H})$
   \STATE expectedCompute, expectedSquaredNorm = compute\_and\_variance$(\bar{D}, \bar{C}, S)$
   \STATE $\eta \leftarrow \bar{\eta} \frac{\text{expectedSquaredNorm}}{\bar{D}_{0, \bar{H}}}$
   \STATE $B += \sum_{i=1}^{\bar{H}} \bar{C}({\bar{H}})$
   \STATE next\_tune $+= \bar{C}({\bar{H}})$
   \ENDIF
   \STATE $N \sim q$
   \FOR{$n=1$ {\bfseries to} $N$}
   \STATE $G_n \leftarrow g(\theta, S[n])$
   \ENDFOR
   \STATE $\hat{G} \leftarrow \sum_{n=1}^N G_n W(n, N)$
   \STATE $\theta \leftarrow \theta - \eta \hat{G}$
   \IF{compute reused}
   \STATE $B += \bar{C}({S[N]})$
   \ELSE
   \STATE $B += \sum_{n=1}^N \bar{C}({S[n]})$
   \ENDIF
   \UNTIL{converged}
\end{algorithmic}
\end{algorithm}

\begin{algorithm}[H]
   \caption{tune}
   \label{alg:tune}
\begin{algorithmic}
   \STATE {\bfseries Input:} current parameter $\theta$,
   current squared distance estimates $\bar{D}_{i,j}$,
   gradient routine $g(\theta, i)$ which returns $\bar{G}_i(\theta)$,
   compute costs $\bar{C}$,
   exponential decay $\alpha$, horizon $\bar{H}$
   \STATE $\bar{G}_0(\theta) \leftarrow 0$
   \FOR{$i=1$ {\bfseries to} $\bar{H}$}
   \STATE $\bar{G}_i(\theta) \leftarrow g(\theta, i)$
   \ENDFOR
   \FOR{$i=0$ {\bfseries to} $\bar{H}$}
   \FOR{$j=1$ {\bfseries to} $\bar{H}$}
   \STATE $D_{i, j} \leftarrow ||G_i - G_j||_2^2$
   \ENDFOR
   \ENDFOR
   \STATE $\bar{D} \leftarrow \alpha \bar{D} + (1 - \alpha) D$
   \STATE $S \leftarrow \text{greedy\_subsequence\_select}(\bar{D}, \bar{C})$
   \STATE $q, W \leftarrow \text{$q$\_and\_$W$}(\bar{D}, \bar{C}, S)$
   \STATE {\bfseries Return:} updated estimates $\bar{D}_{i,j}$,
   sampling distribution $q$, weight function $W$, and subsequence $S$
\end{algorithmic}
\end{algorithm}

\begin{algorithm}[H]
   \caption{greedy\_subsequence\_select}
   \label{alg:greedy}
\begin{algorithmic}
   \STATE {\bfseries Input:} Norm estimates $\bar{D}$, compute costs $\bar{C}$
   \STATE Initialize $N = \text{len}(C)$
   \STATE Initialize $S^+ = [N]$, $S^{-} = [1, ..., N]$,
   converged$ = $FALSE, bestAddCost$ = $cost$(\bar{D}, S^+, \bar{C})$,
   bestRemoveCost$ = $cost$(\bar{D}, S^-, \bar{C})$
   \WHILE{not converged}
   \STATE
   \FOR{$i \in [i $ for $ i \in [1 ... N]$ if not $i \in S^+]$}
   \STATE trial$S \leftarrow $sort$(S^+ + [i])$
   \STATE trialCost$ \leftarrow $cost$(\bar{D}, \bar{C}, $trial$S)$
   \IF{trialCost < bestAddCost}
   \STATE $S^+ \leftarrow $trial$S$
   \STATE bestAddCost$ \leftarrow $trialCost
   \STATE converged $\leftarrow$ False
   \STATE BREAK
   \ELSE
   \STATE converged $\leftarrow$ True
   \ENDIF
   \ENDFOR
   \ENDWHILE
   \STATE converged $\leftarrow$ False
   \WHILE{not converged}
   \FOR{$i \in [i $ for $ i \in S^- $ if$ i \neq N$}
   \STATE trial$S \leftarrow [j $ for $ j \in S^- $if$ j != i]$
   \STATE trialCost$ \leftarrow $sequence\_cost$(\bar{D}, C, trial$S$)$
   \IF{trialCost $<$ bestRemoveCost}
   \STATE $S^- \leftarrow $trial$S$
   \STATE bestRemoveCost$ \leftarrow $trialCost
   \STATE converged $\leftarrow$ False
   \STATE BREAK
   \ELSE
   \STATE converged $\leftarrow$ True
   \ENDIF
   \ENDFOR
   \ENDWHILE
   \IF{bestRemoveCost$ > $ bestAddCost}
   \STATE {\bfseries Return:} $S^-$
   \ELSE
   \STATE {\bfseries Return:} $S^+$
   \ENDIF
\end{algorithmic}
\end{algorithm}

\begin{algorithm}[H]
   \caption{compute\_and\_variance}
   \label{alg:cv}
\begin{algorithmic}
   \STATE {\bfseries Input:} Norm estimates $\bar{D}$, compute costs $\bar{C}$, sequence $S$
   \STATE $q$, $W$ $\leftarrow q$\_and\_$W(\bar{D}, \bar{C}, S)$
   \STATE expectedCompute $\leftarrow \sum_{i \in [1 ... |S|]} q(S[i]]) \bar{C}(S[i])$
   \IF{RT-SS}
   \STATE expectedSquaredNorm $\leftarrow \sum_{i \in [1 ... |S|]} q(S[i]]) W(S[i], S[i]) \bar{D}_{S[i-1], S[i]}$
   \ELSIF{RT-RR}
   \STATE expectedSquaredNorm $\leftarrow \sum_{i \in [1 ... |S|]} \sum_{j \in [1 ... i]} q(S[i]]) W(S[j], S[i]) \bar{D}_{S[j], S[i]}$
   \ELSE
   \STATE Undefined: must specify RT-SS or RT-RR
   \ENDIF
   \STATE {\bfseries Return:} expectedCompute, expectedSquaredNorm
\end{algorithmic}
\end{algorithm}

\begin{algorithm}[H]
   \caption{sequence\_cost}
   \label{alg:cost}
\begin{algorithmic}
   \STATE {\bfseries Input:} Norm estimates $\bar{D}$, compute costs $\bar{C}$, sequence $S$
   \STATE expectedCompute, expectedSquaredNorm = compute\_and\_variance$(\bar{D}, \bar{C}, S)$
   \STATE {\bfseries Return:} expectedCompute * expectedSquaredNorm
\end{algorithmic}
\end{algorithm}

\begin{algorithm}[H]
   \caption{$q$\_and\_$W$}
   \label{alg:qandw}
\begin{algorithmic}
   \STATE {\bfseries Input:} $\bar{D}$, $\bar{C}$, and $S$
   \IF{RT-SS}
   \STATE $q(N) \leftarrow \sqrt{\frac{\bar{D}_{S[N], S[N-1]}}{\bar{C}(S[n])}}$
   \STATE $W(n, N) \leftarrow \frac{1}{q(N)} \mathds{1}\{n=N\}$
   \ELSIF{RT-RR}
   \STATE $\tilde{Q}(N) \leftarrow \sqrt{\frac{\bar{D}_{S[N], S[N-1]}}{\bar{C}(S[n]) - \bar{C}(S[n-1])}}$
   \STATE $\tilde(q)(N) \leftarrow $max$(0, \tilde{Q}(N) - \tilde{Q}(N-1))$
   \STATE $q(N) \leftarrow \frac{\tilde{q}(N)}{\sum_i \tilde{q}(i)}$
   \STATE $W(n, N) \leftarrow \frac{1}{1 - \sum_i q(i)} \mathds{1}\{n \leq N\}$
   \ELSE
   \STATE Undefined: must specify RT-SS or RT-RR
   \ENDIF
   \STATE {\bfseries Return:} $q$, $W$
\end{algorithmic}
\end{algorithm}
\clearpage
\section{Proofs}

\subsection{Proofs for section 2}
\subsubsection{Proposition \ref{prop:unbiased}}
\textbf{Unbiasedness of RT estimators.}
The RT estimators in (\ref{eq:rt_general}) are unbiased estimators of
$Y_H$ as long as
\begin{align*}
\mathbb{E}_{N\sim q} [W(n, N) \mathds{1}\{N \geq n\}] = \sum_{N=n}^H W(n, N)q(N) = 1 \quad \forall n\,.
\end{align*}

\begin{proof}
A randomized telescope estimator which satisfies the above linear constraint condition has expectation:
\begin{align*}
\mathbb{E} [\hat{Y}_H] &= \sum_{N=1}^H q(N) \sum_{n=1}^N W(n, N) \Delta_n\\
&= \sum_{n=1}^H \sum_{N=1}^H \Delta_n W(n, N) q(N)\mathds{1}\{n \leq N\}\\
&= \sum_{n=1}^H \Delta_n \sum_{N=n}^H W(n, N) q(N) = \sum_{n=1}^H \Delta_n = Y_H
\end{align*}
\end{proof}

\subsection{Proofs for section 4}

\subsubsection{Theorem \ref{thm:poly}}
\textbf{Bounded variance and compute with polynomial convergence of $\psi$}.
Assume~$\psi$ converges according to~$\psi_n \leq \frac{c}{(n)^p}$ or faster, for constants ${p > 0}$ and~$c > 0$.
Choose the RT-SS estimator with ${q(n) \propto 1/((n)^{p + 1/2})}$.
The resulting estimator~$\hat{G}$ achieves expected compute~${C \leq (\mathcal{H}_{H}^{p-\frac{1}{2}})^2}$, where~$\mathcal{H}_H^i$ is the~$H$th generalized harmonic number of order~$i$, and expected squared norm~${\mathbb{E}[ ||\hat{G}||_2^2 ] \leq c_{\psi}^2 (\mathcal{H}_H^{p-\frac{1}{2}})^2 := \tilde{G}^2}$.
The limit~${\lim_{H \to \infty} \mathcal{H}_H^{p - \frac{1}{2}}}$ is finite iff~${p > \frac{3}{2}}$, in which case it is given by the Riemannian zeta function,~${\lim_{H \to \infty} \mathcal{H}_H^{p - \frac{1}{2}} = \zeta(p - \frac{1}{2})}$.
Accordingly, the estimator achieves horizon-agnostic variance and expected compute bounds iff~${p > \frac{3}{2}}$.
\begin{proof}
Begin by noting the RT-SS estimator returns $\frac{\Delta_n}{q_n}$ with probability $q(n)$.
Let $\bar{q}(n) = \frac{1}{n^{p + \frac{1}{2}}}$ and
$\sum_{n=1}^H \bar{q}(n) = Z$, such that
$q(n) = \frac{\bar{q}(n)}{Z}$. First, note
$Z = \sum_{n=1}^H \frac{1}{n^{p+\frac{1}{2}}} = \text{H}_{H}^{p+\frac{1}{2}}$.
Now inspect the expected squared norm
$\mathbb{E}||\hat{G}||_2^2$:
\begin{align*}
\sum_{n=1}^H q(n) ||\frac{\Delta_n}{q_n}||_2^2 &=
\sum_{n=1}^H q(n) \frac{||\Delta_n||_2^2}{q_n^2} \\
&= Z \sum_{n=1}^H \bar{q}(n) \frac{||\Delta_n||_2^2}{\bar{q}_n^2}\\
&\leq Z c_\psi^2 \sum_{n=1}^H \bar{q}(n) \frac{n^{2p+1}}{n^{2p}}\\
&= Z c_\psi^2 \sum_{n=1}^H \frac{n^{2p+1}}{n^{3p+\frac{1}{2}}}\\
&= Z c_\psi^2 \sum_{n=1}^H \frac{1}{n^{p-\frac{1}{2}}}\\
&= Z c_\psi^2 \text{H}_H^{p-\frac{1}{2}}\\
&= c_\psi^2 \text{H}_H^{p-\frac{1}{2}} \text{H}_H^{p+\frac{1}{2}}\\
&\leq c_\psi^2 (\text{H}_H^{p-\frac{1}{2}})^2
\end{align*}
Now inspect the expected compute, $\mathbb{E}_{n\sim q} n$:
\begin{align*}
\mathbb{E}_{n\sim q} &= \sum_{n=1}^N q(n) n\\
&= Z \sum_{n=1}^H \frac{n}{n^{p+\frac{1}{2}}}\\
&= Z \sum_{n=1}^H \frac{1}{n^{p-\frac{1}{2}}}\\
&= Z \text{H}_H^{p-\frac{1}{2}}\\
&= \text{H}_H^{p-\frac{1}{2}} \text{H}_H^{p+\frac{1}{2}}\\
&\leq (\text{H}_H^{p-\frac{1}{2}})^2
\end{align*}
\end{proof}

\subsubsection{Theorem \ref{thm:geom}}
\textbf{Bounded variance and compute with geometric convergence of $\psi$}.
Assume~$\psi_n$ converges according to~${\psi_n \leq c p^n}$, or faster, for~${0 < p < 1}$.
Choose RT-SS and with~${q(n) \propto p^n}$.
The resulting estimator ~$\hat{G}$ achieves expected compute~${C \leq (1-p)^{-2}}$ and expected squared norm~${||\hat{G}||_2^2 \leq \frac{c}{(1-p)^2} := \tilde{G}^2}$.
Thus, the estimator achieves horizon-agnostic variance and expected compute bounds for all~${0 < p < 1}$.
\begin{proof}
Let $q(n) = \frac{\bar{q}(n)}{Z}$, for $\bar{q}(n) = p^n$.
Note $Z = \sum_{n=1}^H p^n = p \frac{1 - p^H}{1 - p} \leq \frac{1}{1 - p}$.
Now, note $\psi_n = c_\psi \bar{q}(n)$. It follows
\begin{align*}
\mathbb{E}_{n\sim q} ||\frac{\Delta_n}{q(n)}||_2^2
&= \sum_{n=1}^H q(n) \frac{||\Delta_n||_2^2}{q(n)^2} \\
&\leq\sum_{n=1}^H q(n) \frac{\psi_n^2}{q(n)^2} \\
&=\leq c_\psi^2 \sum_{n=1}^H q(n) \frac{\bar{q}(n)^2}{q(n)^2}\\
&= c_\psi^2 Z^2 \sum_{n=1}^H q(n)\\
&= c_\psi^2 Z^2
\end{align*}
Now consider the expected compute. We have
\begin{align*}
\mathbb{E}_{n\sim q} n &= \sum_{n=1}^N n q(n)\\
&= \sum_{n=1}^N \frac{n p^n}{Z}\\
&= \frac{1}{Z} \sum_{n=1}^N n p^n\\
&= p \frac{1}{Z} \frac{1 + Hp^{H+1} - (H+1)p^H}{(1-p)^2}\\
&= \frac{1 + Hp^{H+1} - (H+1)p^H}{(1-p)(1-p^H)}\\
&\leq \frac{1}{(1-p)(1-p^H)}\\
&\leq \frac{1}{(1-p)^2}
\end{align*}
\end{proof}

\subsubsection{Theorem \ref{thm:infopt}}
\textbf{Asymptotic regret bounds for optimizing infinite-horizon programs}.
Assume the setting from \ref{thm:poly} or \ref{thm:geom}, and the corresponding~$C$ and~$\tilde{G}$ from those theorems.
Let~$R_t$ be the instantaneous regret at the~$t$th step of optimization,~${R_t = \mathcal{L}(\theta_t) - \min_\theta \mathcal{L} (\theta)}$.
Let~$t(B)$ be the greatest~$t$ such that a computational budget~$B$ is not exceeded.
Use online gradient descent with step size~${\eta_t = \frac{D}{\sqrt{t} \mathbb{E}[||\hat{G}||_2^2]}}$.
As~${B \to \infty}$, the asymptotic instantaneous regret is bounded by~${R_{t(B)} \leq \mathcal{O} (\tilde{G} D \sqrt{\frac{C}{B}})}$, independent of~$H$.
\begin{proof}
First, we control $t(B)$ using the central limit theorem.
Note $t \to \infty \iff B(t) \to \infty$. Consider $B$ as a function $B(t)$
of $t$. We have $B(t) = \sum_{\tau=1}^t N_t$, where $N \sim q$. Thus,
$\frac{B(t)}{t} \to \mathbb{E}_{N \sim q} N$ by the central limit theorem. This
implies that in the limit, $t = \frac{B}{C}$.

To complete the proof, plug in $t(B)$ and $\eta_t$, as well as the upper bound
on squared norm $\mathbb{E}||\hat{G}||_2^2 \leq \tilde{G}^2$ and upper bound
on diameter $D$, into standard results for stochastic gradient descent with
convex loss functions (e.g. section 3.4 in \cite{hazan2016introduction})
\end{proof}

\subsection{Proofs for section 5}

\subsubsection{Theorem \ref{thm:advcorr-w}}
\textbf{Optimality of RT-SS under adversarial correlation.}
Consider the family of estimators presented in Equation  \ref{eq:rt_general}.
Assume $\theta$, $\nabla_\theta$, and $G$ are univariate.
For any fixed sampling distribution $q$, the single-sample RT estimator RT-SS minimizes the worst-case variance of $\hat{G}$ across an adversarial choice of covariances~${\Cov(\Delta_i, \Delta_j) \leq \sqrt{\Var(\Delta_i)} \sqrt{\Var(\Delta_j)}}$.

\begin{proof}
Recall $\hat{G} = \sum_{n=0}^N \Delta_n W(n, N)$. Let $\sigma_{i, j}^2 = \Cov(\Delta_i, \Delta_j)$ and $\sigma_i^2 = \Var(\Delta_i)$. The variance of $\hat{G}$ is:
\begin{align*}
\Var(\hat{G}) &= \sum_N q(N) \Big[\sum_{i=0}^N \sum_{j=0}^N W(i, N) W(j, N) \sigma_{i, j}^2 \Big] \\
&\leq \sum_N q(N) \Big[\sum_{i=0}^N \sum_{j=0}^N W(i, N) W(j, N) \sigma_i \sigma_j \Big] \\
&= \sum_N q(N) \Big(\sum_{n=0}^N W(n, N) \sigma_n \Big)^2
\end{align*}
Note the above bound is tight as the adversary can choose $\Cov(\Delta_i, \Delta_j) = \sigma_i \sigma_j$. Introduce $\rho(n, N) = W(n, N) q(N)$, and note that the constraint from proposition \ref{prop:unbiased} can equivalently be stated as $\sum_{N \geq n} \rho(n, N) = 1 \forall n$. We have the variance:
\[
\Var(\hat{G} | N) \leq \sum_N \frac{1}{q(N)} \Big(\sum_{n=0}^N \rho(n, N) \sigma_n \Big)^2
\]
Consider finding $\rho(n, N)$ which minimizes the variance for an arbitrary $q$. The constrained optimization has the Lagrangian:
\[
J = \Big(\sum_N \frac{1}{q(N)} (\sum_{n=0}^N \rho(n, N) \sigma_n )^2\Big) + \sum_n \lambda_n (\sum_{N \geq n} \rho(n, N) - 1)
\]
We can accordingly optimize by taking derivatives:
\begin{align*}
    \frac{dJ}{d\rho(n, N)} &= 2C q(N)(\sum_{i=0}^N w(i, N) \sigma_i) \sigma_n + \lambda_n\\
    \frac{dJ}{d\rho(n, N)} = 0 &\implies \sigma_n q(N) \sum_{i=0}^N w(i, N) \sigma_i = k_n\\
    &\implies \sigma_n \sum_{i=0}^N \rho(i, N) \sigma_i = k_n \forall N \geq n\\
    &\implies \rho(n, N) = 0 \forall N > n
\end{align*}
\end{proof}

\subsubsection{Theorem \ref{thm:advcorr-q}}
\textbf{Optimal q under adversarial correlation.}
Consider the family of estimators presented in Equation  \ref{eq:rt_general}.
Assume $\Cov(\Delta_i, \Delta_i)$ and $\Cov(\Delta_i, \Delta_j)$ are diagonal.
The RT-SS estimator with~${q_n \propto \sqrt{\frac{\mathbb{E}[||\Delta_n||_2^2}{C(n)}}}$ maximizes the ROE across an adversarial choice of diagonal covariance matrices~${\Cov(\Delta_i, \Delta_j)_{kk} \leq \sqrt{\Cov(\Delta_i, \Delta_i)_{kk} \Cov(\Delta_j, \Delta_j)_{kk}}}$.
\begin{proof}
First, note that by the assumption of diagonal covariance between all terms,
the expected squared norm decomposes over indices $k$:
\[
\mathbb{E} ||\hat{G}||_2^2 = \sum_k \mathbb{E} \hat{G}[k]^2
\]
For all choices of $q$, the RT-SS estimator minimizes the worst-case variance and thus
(due to unbiasedness) the expected squared value of each entry in $\hat{G}$.
Because the squared norm decomposes, the RT-SS estimator minimizes the squared
norm for all $q$.

It remains to optimize $q$. We know $\rho(n, N) = 0 \forall N > n$.
Therefore to satisfy the constraint, we have $\rho(N, N) = 1$.
It follows that:
\[
\text{ROE}^-1 = \big( \sum_N q(N) C(N) \big) \big(\sum_N \frac{\mathbb{E}||\Delta_N||_2^2}{q(N)} \big)
\]
We require $\sum_N q(N) = 1$. The constrained optimization has the Lagrangian:
\[
J = \Big( \sum_N q(N) C(N) \Big) \Big(\sum_N \frac{\mathbb{E}||\Delta_N||_2^2}{q(N)}\Big) + \lambda (\sum_N q(N) - 1)
\]
Let $C = \Big( \sum_N q(N) C(N) \Big)$ and $V = \Big(\sum_N \frac{\mathbb{E}||\Delta_N||_2^2}{q(N)}\Big)$. We optimize $q(N)$ by taking the derivative of the inverse ROE:
\begin{align*}
\frac{d\text{ROE}^{-1}}{dq(N)} &= C(N) V - C \frac{\sigma_N^2}{q(N)^2}\\
\frac{d\text{ROE}^{-1}}{dq(N)} = 0 &\implies q(N)^2 \propto \frac{\mathbb{E}||\Delta_N||_2^2 C}{C(N) V}\\
&\implies q(N) \propto \sqrt{\frac{\mathbb{E}||\Delta_N||^2_2}{C(N)}}
\end{align*}
\end{proof}

\subsubsection{Theorem \ref{thm:nocorr-w}}
\textbf{Optimality of RT-RR under independence}.
Consider the family of estimators presented in Eq.~\ref{eq:rt_general}.
Assume the $\Delta_j$ are univariate.
When the~$\Delta_j$ are uncorrelated, for any importance sampling distribution~$q$, the Russian roulette estimator achieves the minimum variance in this family and thus maximizes the optimization efficiency lower bound.
\begin{proof}
By independence, we have $\mathbb{E}\big(\sum_n W(n, N) \Delta_n \big)^2 = \sum_n W(n, N)^2 \mathbb{E} \Delta_n^2$. It follows that an RT estimator has variance:
\begin{align*}
\Var(\hat{G}) &= \sum_N q(N) \sum_{n\leq N} W(n, N)^2 \mathbb{E} \Delta_n^2\\
&= \sum_N \frac{1}{q(N)} \sum_{n\leq N} \rho(n, N)^2 \mathbb{E} \Delta_n^2
\end{align*}
Recall the constraint in proposition \ref{prop:unbiased} requires $\sum_{N \geq n} \rho(n, N) = 1$ for all $n$. The Lagrangian of the constrained minimization of $\Var(\hat{G})$ with respect to $\rho$ is:
\[
J = \Var(\hat{G}) + \sum_n \lambda_n (\sum_{N \geq n} \rho_n - 1)
\]
We optimize $\rho$ by finding the minimum of the Lagrangian:
\begin{align*}
\frac{dJ}{d\rho(n,N)} &= \frac{2}{q(N)} \rho(n, N) \mathbb{E} \Delta_n^2 + \lambda_n \\
\frac{dJ}{d\rho(n,N)} = 0 &\implies \frac{\rho(n, N)}{q(N)} = -\frac{\lambda_n}{2 \mathbb{E} \Delta_n^2}\\
&\implies W(n, N) = -\frac{\lambda_n}{2 \mathbb{E} \Delta_n^2}, \text{ which is independent of } N\\
&\implies W(n, N) = \frac{1}{\sum_{N' \geq n} q(N')} \text{ to fulfill the constraint in proposition }\ref{prop:unbiased}
\end{align*}
\end{proof}

\subsubsection{Theorem \ref{thm:nocorr-q}}
\textbf{Optimal q under independence}.
Consider the family of estimators presented in Equation  \ref{eq:rt_general}.
Assume $\Cov(\Delta_i, \Delta_i)$ is diagonal and~$\Delta_i$ and~$\Delta_j$ are independent.
The RT-RR estimator with
$Q(i) \propto \sqrt{\frac{\mathbb{E} [||\Delta_i||_2^2}{C(i) - C(i-1)}]}$,
where $Q(i) = \Pr(n \geq i) = \sum_{j=i}^H q(j)$,
maximizes the ROE.

\begin{proof}
First note that by theorem \ref{thm:nocorr-w}, for any $q$ and for each element in the vector
$\hat{G}$, the RT-RR estimator minimizes the variance of that element. Now note
that due to independence of $\Delta_i, \Delta_j$ and diagonality of $\Cov(\Delta_i, \Delta_i)$:
\begin{align*}
\mathbb{E} ||\sum_{n=1}^N W(n, N) \Delta_n||_2^2 &= \sum_{n=1}^N W(n, N) \mathbb{E} ||\Delta_n||_2^2\\
&= \sum_k \sum_{n=1}^N W(n, N) \mathbb{E}\Delta_n[k]^2
&= \sum_k \mathbb{E} \hat{G}[k]^2
\end{align*}
As the RT-RR estimator minimizes $\mathbb{E} \hat{G}[k]^2$ for each coordinate $k$,
it also minimizes $\mathbb{E} ||\hat{G}||_2^2$. It remains to optimize $Q$.
Consider the inverse ROE of the RT-RR estimator. By independence we have:
\begin{align*}
    \text{ROE}(\hat{G})^{-1} = \mathbb{E} ||\hat{G}||_2^2 \mathbb{E} C &=
    \Big(\sum_N q(N) \sum_{n\leq N} \frac{1}{Q(n)^2} \mathbb{E} ||\Delta_n||_2^2\Big) \Big(\sum_N q(N) C(N)\Big)
\end{align*}
Take the gradient of the inverse optimization efficiency lower bound w.r.t. $q(n)$:
\[
\frac{d\text{ROE}(\hat{G})^{-1}}{dq(N)} = C(N) \mathbb{E} ||\hat{G}||_2^2 + \sum_{n \leq N} \frac{1}{Q(n)^2} \mathbb{E} ||\Delta_n||_2^2 - \sum_{i} q(i) \sum_{j \leq min(i, N)} \frac{2}{Q(j)^3} \mathbb{E} ||\Delta_j||_2^2\\
\]
\begin{align*}
\sum_{i} q(i) \sum_{j \leq min(i, N)} \frac{2}{Q(j)^3} \mathbb{E} ||\Delta_j||_2^2 &= \sum_{j \leq N} \frac{2}{Q(j)^2} \mathbb{E} ||\Delta_j||_2^2 \frac{\sum_i q(i) \mathds{1}\{i \geq j\}}{Q(j)}\\
&= \sum_{j \leq N} \frac{2}{Q(j)^2} \mathbb{E} ||\Delta_j||_2^2 \quad \text{ by definition of } Q(j)
\end{align*}
\[
\implies \frac{d\text{ROE}(\hat{G})^{-1}}{dq(N)} = C(N) \mathbb{E} ||\hat{G}||_2^2 - \sum_{n \leq N} \frac{1}{Q(n)^2} \mathbb{E} ||\Delta_n||_2^2
\]
Now optimize the objective w.r.t. $Q$ by finding the critical point:
\begin{align*}
\frac{d\text{ROE}(\hat{G})^{-1}}{dq(N)} = 0 \implies C(N) \mathbb{E} ||\hat{G}||_2^2 &=  \sum_{n \leq N} \frac{1}{Q(n)^2} \mathbb{E} ||\Delta_n||_2^2\\
\implies \mathbb{E} ||\hat{G}||_2^2 \Big(C(N) - C(N-1)\Big) &= \frac{1}{2}  \frac{\mathbb{E}||\Delta_N||_2^2}{Q(N)^2}\\
\implies Q(N)^2 &\propto \frac{\mathbb{E} ||\Delta_n||_2^2}{C(N) - C(N-1)}
\end{align*}
\end{proof}
\end{document}